\documentclass[lettersize,journal]{IEEEtran}
\usepackage{amsmath,amsfonts}
\usepackage{algorithmic}
\usepackage{array}
\usepackage[
    caption=false,
    font=normalsize,
    labelfont=rm,
    textfont=rm
]{subfig}\usepackage{textcomp}
\usepackage{stfloats}
\usepackage{url}
\usepackage{verbatim}
\usepackage{graphicx}

\usepackage{booktabs}
\usepackage{multirow}
\usepackage[hidelinks]{hyperref}

\def\BibTeX{{\rm B\kern-.05em{\sc i\kern-.025em b}\kern-.08em
    T\kern-.1667em\lower.7ex\hbox{E}\kern-.125emX}}
\usepackage{balance}

\begin{document}

\title{LINK: Lightweight and Inductive Node-to-Node Propagation Rule Learning for Traffic Kriging}

\author{
You~Shang,
Yu-Kwan~Ching,
Weifeng~Su,
Guanyao~Li,
Ruiyuan~Li,
and Gary~Shueng-Han~Chan

\thanks{
You Shang is with Beijing Normal-Hong Kong Baptist University,
Zhuhai, China, and Hong Kong Baptist University,
Hong Kong SAR, China.
E-mail: shangyou@bnbu.edu.cn.
}
\thanks{
Weifeng Su and Guanyao Li are with
Beijing Normal-Hong Kong Baptist University,
Zhuhai, China.
E-mail: \{wfsu, guanyaoli\}@bnbu.edu.cn.
}%
\thanks{
Yu-Kwan Ching and Gary Shueng-Han Chan are with
The Hong Kong University of Science and Technology,
Hong Kong SAR, China.
E-mail: ykching@connect.ust.hk;
gchan@cse.ust.hk.
}%
\thanks{
Ruiyuan Li is with Chongqing University,
Chongqing, China.
E-mail: ruiyuan.li@cqu.edu.cn.
}%
\thanks{
Corresponding author: Guanyao Li
(guanyaoli@bnbu.edu.cn).
}%
\thanks{Our code is publicly available in https://github.com/AlbertShang/LINK}
}

\markboth{Journal of \LaTeX\ Class Files,~Vol.~18, No.~9, September~2020}%
{How to Use the IEEEtran \LaTeX \ Templates}

\maketitle

\begin{abstract}

Spatio-temporal traffic kriging aims to reconstruct traffic time series at unobserved locations with no historical traffic observations by leveraging spatial and temporal dependencies across the road network.
Existing Graph Neural Network (GNN)-based approaches primarily rely on neighborhood aggregation to capture spatial dependencies, thereby modeling traffic interactions implicitly and making it difficult to accurately recover traffic magnitudes at locations without historical observations. To address this limitation, we propose LINK, a Lightweight and Inductive Node-to-node propagation rule learning paradigm for traffic Kriging. 
LINK introduces three key designs: 1) a propagation-centric traffic kriging paradigm that explicitly learns a shared node-to-node transition rule; 2) a context-aware Spectral Propagation Network that performs traffic-state propagation in both the spectral and time domains; and 3) TraffiKrig, a realistic traffic kriging benchmark that preserves heterogeneous road-segment characteristics and road-specific metadata while enabling systematic comparison of representative kriging methods under a unified metadata-enhanced setting.
Extensive experiments demonstrate that LINK achieves strong reconstruction accuracy and computational efficiency.
Compared with the strongest baseline on each subset, LINK achieves an average MAE reduction of \textbf{54.7\%} over original baselines. 
LINK also achieves substantial efficiency improvements, reducing training time by \textbf{66.5\%} and inference time by \textbf{86.6\%} compared with the fastest competing methods.
Zero-shot experiments further demonstrate the transferability of the learned propagation rule across regions.
\end{abstract}

\begin{IEEEkeywords}
Spatio-temporal data mining, traffic kriging, inductive representation learning.
\end{IEEEkeywords}

\section{Introduction}\label{sec:intro}

Accurate and comprehensive traffic state information is fundamental to many intelligent transportation applications, including traffic monitoring~\cite{kamijo2000monitoring}, congestion management~\cite{wang2024congestion}, route planning~\cite{xu2023routing}, and urban mobility analysis~\cite{wang2022mobility1,senaratne2017mobility2}. In practice, however, traffic sensors can only be deployed at a limited number of road segments or intersections because of installation, communication, and maintenance costs~\cite{rodriguez2019sparse1,shan2015sparse2}. As a result, traffic states are continuously available only at a subset of locations, while a large portion of the road network remains unobserved, limiting downstream applications that require comprehensive traffic information. Traffic kriging is commonly used to address this problem.


Given a road network where only a subset of nodes have historical traffic measurements, traffic kriging aims to recover the traffic series at unobserved nodes, which have no historical traffic observations of their own, using the measurements from observed nodes together with road-network structure and auxiliary road information~\cite{wu2021IGNNK}.


Traffic kriging is fundamentally different from conventional traffic forecasting~\cite{ma2025forecasting2,Duc2025forecasting1,li2022forecasting3} and traffic data imputation~\cite{duan2016imputation}. In traffic forecasting, historical observations at the target location are available to predict its future states. Likewise, traffic imputation typically reconstructs missing values from a partially observed temporal sequence at the target location. In contrast, traffic kriging primarily relies on transferring information from observed locations to the target locations without historical observations through the road network, giving rise to following three intrinsic challenges.

First, informative observed nodes may lie at substantially different graph distances for unobserved targets. This heterogeneous support structure makes it difficult to characterize spatial dependencies with a fixed propagation range, since the amount and distance of useful information vary substantially across targets.
Second, the transformation of a traffic state between two road-network nodes is not uniform, but depends on their source--destination relationship, road characteristics, local network structure, and temporal context~\cite{jiang2023context1,guo2021astgnn}. This heterogeneity makes it difficult to characterize a transition mechanism that can adapt to diverse road contexts while remaining generalizable across different locations.
Third, the traffic state being transferred contains heterogeneous temporal dynamics and cannot be equally transferable across locations, making it difficult for a single representation to preserve different temporal characteristics.

Existing traffic kriging methods predominantly use graph neural networks (GNNs) to model dependencies over road networks~\cite{xu2025kits,wei2024iagcn,li2025stagann}. Although effective, their layer-wise representation aggregation paradigm leaves two important issues unresolved for cross-location traffic-state transfer. First, GNN-based methods mainly update node representations through layer-wise neighborhood aggregation, leaving the transformation of traffic states between connected locations implicit. Second, all nodes in existing works undergo the same number of message-passing rounds to propagate information~\cite{liu2022gnnsurvey}. Such fixed-depth propagation is less suitable for traffic kriging, where effective observational support varies across targets and propagation branches.

To tackle the challenges and limitations of existing work, we propose \textbf{LINK}, a \textbf{L}ightweight and \textbf{I}nductive \textbf{N}ode-to-node propagation rule learning paradigm for traffic \textbf{K}riging. LINK shifts the modeling focus from node representation reconstruction to explicit and transferable traffic state propagation. 

For each target node, LINK first constructs a focal-specific propagation subgraph that determines where information is drawn from. Different focal nodes, and different branches of the same focal node, can therefore naturally have different propagation depths. This allows LINK to preserve the effective observational support of each target without unnecessarily expanding propagation over the entire graph.
LINK then models how traffic states are transferred through a shared node-to-node propagation rule. The same rule is repeatedly applied across different road-network edges, propagation hops, and focal-specific subgraphs. LINK thus generalizes a shared traffic transition mechanism rather than representations tied to particular node identities. Road-specific metadata, pairwise road relations, local structural information, and temporal context are further incorporated as transition context, allowing the shared rule to adapt to heterogeneous road-network conditions.

Furthermore, we instantiate LINK with a Spectral Propagation Network (SPN), which decomposes each traffic series into a slowly varying spectral trend and a temporal residual.
The trend component is represented and propagated in the frequency domain through a shared spectral propagation rule, while the residual component is propagated and refined in the time domain to recover short-term variations.
This dual-domain design enables traffic structures at different temporal scales to be modeled with complementary representations, rather than forcing them into a single propagation space.

To enable traffic kriging to be evaluated under realistic road-network contexts, we further construct \textbf{TraffiKrig}, a benchmark derived from TraffiDent~\cite{gou2026traffident}.
Unlike commonly used traffic benchmarks that provide limited road-context information, TraffiKrig preserves the directed road-network topology together with road-specific metadata, including road type, travel direction, and freeway information.
Extensive experiments on TraffiKrig demonstrate that LINK achieves both strong reconstruction accuracy and computational efficiency.
Compared with the strongest baseline evaluated in its original setting on each subset, LINK achieves an average MAE reduction of \textbf{54.7\%}. It is also significantly more computationally efficient, reducing training time by \textbf{66.5\%} and inference time by \textbf{86.6\%} compared with the fastest baseline. 
Moreover, zero-shot experiments demonstrate that LINK achieves competitive performance on unseen regions, providing further evidence that it learns a transferable propagation rule. Meanwhile, several metadata-enhanced baselines achieve MAE reductions of over \textbf{40\%} compared with their original counterparts, highlighting the value of road context for traffic kriging.


Our main contributions are summarized as follows:
\begin{itemize}
    \item \textbf{Propagation-Centric Traffic Kriging.}
    We formulate traffic kriging as a transferable traffic state propagation problem and introduce LINK, a lightweight and inductive paradigm that explicitly learns a shared node-to-node transition rule instead of relying solely on implicit representation aggregation. The learned rule is shared across source--destination pairs and propagation depths, and can be directly applied to all propagation edges.

    \item \textbf{Shared Context-Aware Dual-Domain Propagation.}

    We instantiate LINK with a Spectral Propagation Network (SPN), which learns transferable propagation rules conditioned on road context. SPN is shared across node-to-node transitions and propagation depths, enabling an efficient and scalable kriging process.
    To capture heterogeneous temporal dynamics, SPN propagates dominant temporal structures in the frequency domain while refining short-term variations in the time domain.

    \item \textbf{Realistic Benchmark and Comprehensive Evaluation.}
We construct TraffiKrig, a traffic kriging benchmark that preserves heterogeneous road types and road-specific metadata while supporting a unified metadata-enhanced evaluation protocol for systematic comparison across representative kriging methods.
    Comprehensive experiments demonstrate the effectiveness and computational efficiency of LINK, and reveal the substantial value of road metadata for traffic kriging.
\end{itemize}

The remainder of this paper is organized as follows. We first discuss related works in Section~\ref{sec:related}. After preliminaries in Section~\ref{sec:preliminary}, we detail LINK in Section~\ref{sec:methodology}. We present the experimental settings and results in Section~\ref{sec:exp}, and conclude in Section~\ref{sec:conclusion}.

\section{Related Work}\label{sec:related}

\subsection{Graph Neural Network }

Graph Neural Networks (GNNs) have been widely adopted for kriging tasks, typically learning node representations by aggregating neighborhood information  ~\cite{jin2024survey}. Classical models such as GCN~\cite{kipf2016gcn}, GAT~\cite{veličković2018gat} and GIN~\cite{xu2018gin} rely on neighborhood aggregation, while GraphSAGE~\cite{hamilton2017graphsage} enables inductive representation learning through neighborhood aggregation functions shared across nodes within each layer. These methods exploit inter-node relationships primarily for learning node representations. However, traffic systems exhibit highly dynamic and fluctuating behaviors with complex spatio-temporal dependencies~\cite{zhang2024imputationsurvey,wu2019gwn,li2018dcrnn}, which cannot be fully characterized by simply aggregating neighborhood information without explicitly modeling state transitions between sensors. To address this limitation, LINK explicitly models inter-node state transitions through a spectral propagation network. 
Unlike conventional GNNs that adopt a fixed propagation depth for all nodes~\cite{gilmer2017mpnn,zeng2021fixdepth}, where a $K$-layer GNN uniformly propagates information across $K$ hops for every node~\cite{liu2022gnnsurvey}, LINK performs propagation on focal-specific adaptive subgraphs. By tailoring the propagation field to each target location, LINK avoids unnecessary aggregation from irrelevant nodes and improves both reconstruction accuracy and computational efficiency.
This distinction in propagation strategy between LINK and conventional GNN message passing is further illustrated through a concrete example in Section~\ref{sec:propagation}.

\subsection{Spatio-Temporal Kriging}

Spatio-temporal kriging incorporates the temporal dimension into spatial kriging, aiming to recover time-series signals at unobserved locations from observations at sampled locations~\cite{wu2021IGNNK}. Early spatio-temporal kriging methods mainly rely on statistical models to characterize spatial dependencies~\cite{wikle2015statickriging1,guttorp2013statickriging2}. Gaussian process-based approaches model the correlations among locations through covariance functions, assuming that spatial dependencies can be captured by geometric proximity and smoothness~\cite{zhan2025gp4kriging}.

With the emergence of GNNs, recent studies have explored graph-based approaches for spatio-temporal kriging, where spatial dependencies are modeled through graph structures.

Existing inductive kriging methods mainly focus on improving the generalization from observed sensors to unseen sensors. IGNNK~\cite{wu2021IGNNK} introduces a masking-based training strategy, where observed nodes are randomly masked and reconstructed to mimic unseen sensors during inference. KITS~\cite{xu2025kits} further reduces the train-inference graph gap by inserting virtual nodes during training to mimic unseen sensors. 

Another line of research focuses on improving the quality of spatial relationships, as predefined graphs may not sufficiently characterize the complex dependencies among sensors. IAGCN~\cite{wei2024iagcn} and STA-GANN~\cite{li2025stagann} argue that fixed graph structures cannot fully capture complex spatial correlations among sensors. They therefore construct adaptive or dynamic graphs to learn complex spatial relationships from traffic data and auxiliary information. INCREASE~\cite{zheng2023increase} leverages points of interest and functional areas to incorporate functional information, whereas SATCN~\cite{wu2021satcn} leverages K-nearest neighbors and pairwise distances to incorporate geographic proximity information. Moreover, GenCast~\cite{su2026gencast} extends auxiliary-information-based kriging by incorporating external environmental signals, such as weather data, to enhance generalization across regions.

However, existing approaches achieve inductive generalization mainly through auxiliary training mechanisms, which introduce additional complexity into the training process. Meanwhile, efforts to improve spatial modeling primarily focus on refining graph connectivity or spatial correlations, while the underlying traffic propagation process between sensors remains implicit. 
In contrast, LINK performs training directly on focal-specific adaptive subgraphs, avoiding the additional computational overhead introduced by auxiliary training strategies. Furthermore, LINK explicitly learns transferable traffic state propagation rules and reuses them across different focal nodes during inference, improving both inference efficiency and the accuracy of node-to-node transition modeling.




\section{Preliminary}
\label{sec:preliminary}

This section introduces the fundamental concepts and notations used throughout this paper. We first formulate the traffic kriging problem over road networks, including the road-network representation and the observed/unobserved node setting. We then briefly review the Discrete Fourier Transform (DFT) and the real-valued Fast Fourier Transform (rFFT), which form the spectral basis of the proposed Spectral Propagation Network.

\subsection{Problem Definition}

Traffic dynamics are closely associated with the underlying road network, where road-specific meta-features provide vital information for characterizing traffic states~\cite{gou2026traffident}. Accordingly, we study spatio-temporal kriging over road networks by explicitly considering road meta-features, which we refer to as traffic kriging in this work.

\textbf{Road Network.}
We abstract the physical road network as a graph $\mathcal{G}=(\mathcal{V},\mathcal{E},\mathbf{A},\mathbf{M})$, where $\mathcal{V}$ denotes a set of $|\mathcal{V}|=N$ road-network nodes, each corresponding to a road-network unit such as a road segment or an intersection, and $\mathcal{E}$ denotes the topological connections among them. Nodes with available sensor measurements provide observable traffic states, while nodes without available observations are treated as unobserved locations  whose traffic states need to be inferred. $\mathbf{A}\in\mathbb{R}^{N\times N}$ denotes the adjacency matrix that characterizes the spatial connectivity. $\mathbf{M}=\{\mathbf{m}_i\}_{i=1}^{N}$ denotes the static road metadata, where $\mathbf{m}_i$ contains road-specific contextual information associated with node $v_i$. The traffic states over $T$ time steps are represented by $\mathbf{X}\in\mathbb{R}^{N\times T}$. For node $v_i$, its traffic series is denoted by $\mathbf{x}_{i}=[x^1_{i},x^2_{i},\ldots,x^T_{i}]$, where $x_{i}^t$ represents the traffic state at node $v_i$ and time step $t$.

\textbf{Traffic Kriging.}
According to the availability of traffic observations, we partition the road-network nodes into an observed set $\mathcal{V}_{o}$ and an unobserved set $\mathcal{V}_{u}$. Nodes in $\mathcal{V}_{o}$ are associated with available sensor measurements, whereas nodes in $\mathcal{V}_{u}$ do not have available traffic observations. For each node $v_i\in\mathcal{V}_{u}$, its entire traffic series $\mathbf{x}_{i}$ is unavailable. Let $\mathbf{X}_{o}$ and $\mathbf{X}_{u}$ denote the traffic series corresponding to $\mathcal{V}_{o}$ and $\mathcal{V}_{u}$, respectively. Traffic kriging aims to learn a function $\mathcal{F}$ that estimates the traffic series at unobserved road-network nodes from the available traffic observations, road-network topology, and road meta-features:
\begin{equation}
    \hat{\mathbf{X}}_{u}
    =
    \mathcal{F}
    \left(
        \mathbf{X}_{o},
        \mathcal{E},
        \mathbf{A},
        \mathbf{M}
    \right).
\end{equation}

\subsection{Fourier Transform}
Frequency-domain representations have been increasingly explored in time-series modeling, as they enable temporal patterns at different frequencies to be explicitly characterized~\cite{yi2023frequency1,xu2024frequency2}. 
The Fourier transform (FT) represents a temporal signal in the frequency domain. For discrete temporal sequences, its discrete counterpart is the Discrete Fourier Transform (DFT), while the Fast Fourier Transform (FFT) provides an efficient algorithm for computing the DFT~\cite{fourier}. Since traffic series are real-valued, we employ the real-valued Fast Fourier Transform (rFFT), which only retains the non-redundant frequency components.

For a road-network node $v_i$, given a temporal window $\mathbf{x}_{i}^{t,W}=[x_i^{t-W+1},\ldots,x_i^t]$ of length $W$, its $k$-th rFFT coefficient is defined as
\begin{equation}
    z_{i,k}
    =
    \sum_{n=0}^{W-1}
    x_i^{t-W+1+n}
    e^{-j2\pi kn/W},
    \qquad
    k=0,\ldots,\left\lfloor\frac{W}{2}\right\rfloor,
\end{equation}
where $j$ denotes the imaginary unit and $z_{i,k}$ denotes the complex-valued coefficient of the $k$-th frequency component.

Each coefficient can be expressed as
\begin{equation}
    z_{i,k}
    =
    A_{i,k}e^{j\phi_{i,k}},
\end{equation}
where $A_{i,k}=|z_{i,k}|$ and $\phi_{i,k}=\arg(z_{i,k})$ denote its amplitude and phase, respectively. From a representation perspective, the frequency index $k$ characterizes the temporal scale of variation, the amplitude $A_{i,k}$ reflects the strength of the corresponding temporal pattern, and the phase $\phi_{i,k}$ describes its temporal alignment. Low-frequency components generally capture slowly varying temporal structures, whereas higher-frequency components characterize faster variations. The corresponding temporal signal can be reconstructed from the rFFT coefficients through the inverse real-valued Fast Fourier Transform (irFFT).

\section{Methodology}\label{sec:methodology}

To explicitly model traffic propagation under road-network constraints, we first introduce the LINK paradigm for traffic kriging in Section~\ref{sec:adaptive-paradigm}. LINK defines focal-specific propagation structures and a transferable propagation rule that can be repeatedly applied across different subgraphs and propagation depths. We then instantiate this paradigm with a concrete traffic kriging model, SPN, and describe its components and implementation details in Section~\ref{sec:model}.

\begin{figure*}[t]
    \centering
    \includegraphics[width=\linewidth]{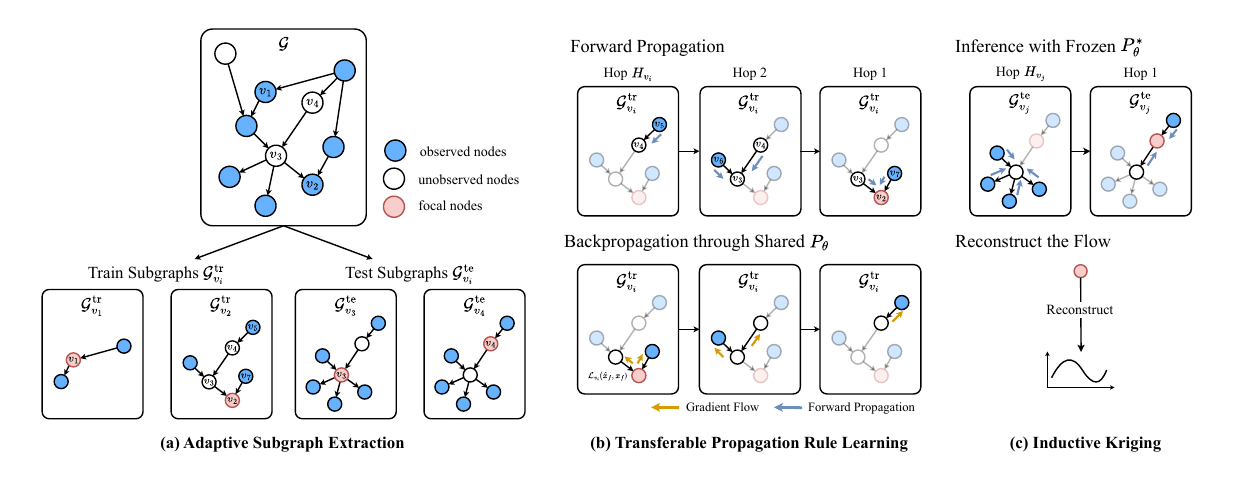}
    \caption{Overview of LINK.}
    \label{fig:paradigm}
\end{figure*}

 \subsection{Node-to-Node Propagation Rule Learning}
\label{sec:adaptive-paradigm}

As illustrated in Fig.~\ref{fig:paradigm}, the LINK paradigm consists of three stages. Adaptive Subgraph Extraction constructs a focal-specific propagation subgraph for each target node. Shared Propagation Rule Learning learns a transferable node-to-node propagation rule $P_{\theta}^{*}$ that can be reused across different edges, propagation depths, and focal-specific subgraphs. Inductive Imputation applies the learned propagation rule to focal-specific test subgraphs to recover the traffic series of unobserved nodes.

\subsubsection{Adaptive Subgraph Extraction}\label{sec:subgraph}

In traffic networks, the traffic state at a road-network node can be influenced by traffic propagated from its surrounding nodes. Adaptive subgraph extraction aims to identify a set of surrounding nodes that provide effective propagation support for each focal node from the complete road-network graph $\mathcal{G}$, where a focal node denotes a target node $v$ whose traffic states are to be imputed.

As illustrated in Fig.~\ref{fig:paradigm}(a), the blue nodes represent observed nodes $\mathcal{V}_o$, while the white nodes represent unobserved nodes $\mathcal{V}_u$ whose road-network structure is known but traffic states are unavailable. Given a focal node, its focal-specific subgraph is extracted by searching along the road network for a set of observed boundary nodes that provide propagation support for the focal node. Specifically, we perform breadth-first search (BFS) from the focal node along the road-network topology. When an unobserved node is reached, the search continues outward along that branch; when an observed node is reached, it is included as a boundary node and the search along that branch terminates. Since the road network is directed, the search is conducted separately along both upstream and downstream directions. After assigning hop distances by BFS, we retain only propagation edges that move one hop closer to the focal node, ensuring that traffic states from observed boundary nodes are propagated toward the focal node along shortest-hop paths. The resulting subgraph therefore contains the focal node, intermediate unobserved nodes, and observed boundary nodes connected by propagation edges directed toward the focal node.

During training, we traverse the observed nodes and treat each of them as a pseudo-focal node by temporarily withholding its own traffic states. We thereby construct a set of focal-specific training subgraphs $\{\mathcal{G}^{\mathrm{tr}}_{v_i}\mid v_i\in\mathcal{V}_o\}$, which are used to learn the shared propagation rule. During inference, we similarly construct the test subgraphs $\{\mathcal{G}^{\mathrm{te}}_{v_j}\mid v_j\in\mathcal{V}_u\}$, on which the learned propagation rule is applied to perform inductive imputation. As illustrated in Fig.~\ref{fig:paradigm}(a), nodes $v_1$ and $v_2$ are sequentially treated as pseudo-focal nodes to construct the training subgraphs, while $v_3$ and $v_4$ are treated as focal nodes to construct the test subgraphs. Taking the training subgraph $\mathcal{G}_{v_2}^{\mathrm{tr}}$ as an example, BFS starts from the focal node $v_2$ and searches outward along the road-network topology. Along one branch, the observed node $v_7$ is reached directly and therefore serves as a boundary node. Along another branch, the search first reaches the unobserved node $v_3$ and then continues through $v_4$ until the observed node $v_5$ is reached. The remaining branch follows the same search procedure. Once the nearest observed node has been reached along each branch, the search on that branch terminates, and the resulting nodes and edges form the focal-specific training subgraph $\mathcal{G}_{v_2}^{\mathrm{tr}}$.

\subsubsection{Transferable Propagation Rule Learning}\label{sec:propagation}

To shift the modeling focus from node representation learning to traffic propagation modeling, we design a transferable propagation rule that explicitly characterizes how traffic states are transferred between connected nodes. As illustrated in Fig.~\ref{fig:paradigm}(b), the transferable propagation rule is learned through hop-by-hop forward propagation over the training subgraphs and end-to-end backpropagation from the focal nodes over all focal-specific training subgraphs $\{\mathcal{G}_{v_i}^{\mathrm{tr}}\mid v_i\in\mathcal{V}_o\}$.

\textbf{Forward Propagation.} During forward propagation, we learn an edge-wise propagation rule $P_{\theta}$ from the focal-specific training subgraphs. For each training subgraph $\mathcal{G}_{v_i}^{\mathrm{tr}}$, $P_{\theta}$ is applied to every directed propagation edge to model the state transition between two connected nodes. The same $P_{\theta}$ is shared across all training subgraphs, propagation directions, edges, and propagation hops. As indicated by the blue arrows in Fig.~\ref{fig:paradigm}(b), traffic states are propagated hop by hop from the observed boundary nodes toward the focal node. By repeatedly applying the shared propagation rule, information originating from boundary nodes at different propagation depths can be progressively transferred to the focal node. For each training subgraph $\mathcal{G}_{v_i}^{\mathrm{tr}}$, we denote its maximum propagation depth by $H_{v_i}$. Forward propagation starts from hop $H_{v_i}$ and proceeds hop by hop toward the focal node. Since the spatial distribution of observed and unobserved nodes varies across focal nodes, the maximum propagation depth $H_{v_i}$ varies across training subgraphs and thus require different numbers of propagation steps.

When multiple propagated states arrive at the same node, an aggregation operator is used to combine them. The proposed paradigm is compatible with different aggregation functions, such as mean, sum, or adjacency-weighted aggregation. In our implementation, we adopt mean aggregation, since the corresponding edge weight is already incorporated into the input features of $P_{\theta}$. Only valid propagated states participate in aggregation. If no valid message reaches a node at a propagation hop, the node is marked as invalid and does not contribute to subsequent propagation. At the focal node, the upstream and downstream estimates are averaged when both are valid; otherwise, the valid directional estimate is used directly.

Fig.~\ref{fig:paradigm}(b) illustrates the propagation process on the training subgraph $\mathcal{G}_{v_i}^{\mathrm{tr}}$ with $i=2$. The blue arrows indicate applications of the shared propagation rule $P_{\theta}$. At Hop 3, the traffic state of the farthest boundary node $v_5$ is first propagated to $v_4$. At Hop 2, the same $P_{\theta}$ is applied to propagate the state of the observed boundary node $v_6$ and the propagated state of $v_4$ obtained from the previous hop toward $v_3$, where the resulting states are aggregated by mean. At Hop 1, the same procedure is applied to propagate the states from $v_3$ and $v_7$ toward the focal node $v_2$, completing the forward propagation process on $\mathcal{G}_{v_2}^{\mathrm{tr}}$.

This propagation process is fundamentally different from conventional GNN--based method. In the example above, allowing the information from $v_5$, which is three hops away, to reach the focal node $v_2$ would require at least three message-passing layers in a standard one-hop GNN. Consequently, although the information from the nearby observed node $v_7$ can reach $v_2$ after only one hop, it continues to be transformed during the subsequent message-passing layers. In contrast, LINK follows the focal-specific propagation structure and applies the shared propagation rule according to the propagation depth of each branch, avoiding unnecessary repeated propagation of nearby observations. 


\textbf{Backpropagation through Shared $P_\theta$.} After the complete multi-hop propagation process, the prediction at the focal node is compared with its ground-truth traffic state to compute the loss $\mathcal{L}_{v_i}(\hat{\mathbf{x}}_{v_i},\mathbf{x}_{v_i})$. A single backward pass is then performed through the entire unrolled propagation process. Since the same $P_{\theta}$ is repeatedly applied across different edges and propagation hops, the gradient with respect to its shared parameters is accumulated over all applications:
\begin{equation}
\nabla_{\theta}
\mathcal{L}_{v_i}
\bigl(\hat{\mathbf{x}}_{v_i},\mathbf{x}_{v_i}\bigr)
=
\sum_{c\in\mathcal{C}(v_i)}
\left[
\nabla_{\theta}
\mathcal{L}_{v_i}
\bigl(\hat{\mathbf{x}}_{v_i},\mathbf{x}_{v_i}\bigr)
\right]_{c},
\end{equation}
where $\mathcal{C}(v_i)$ denotes all applications of the shared propagation rule in the computation graph associated with focal node $v_i$, and $\left[
\nabla_{\theta}
\mathcal{L}_{v_i}
(\hat{\mathbf{x}}_{v_i},\mathbf{x}_{v_i})
\right]_{c}$ denotes the gradient contribution propagated through the $c$-th
application of $P_{\theta}$. Therefore, gradient contributions from different propagation edges and hops are accumulated to update the shared parameters $\theta$.

The aggregation operators between successive propagation hops are also part of the differentiable computation graph. During backpropagation, the gradient at a destination node is distributed to its valid incoming propagated states according to the aggregation rule. Under the mean aggregation used in our implementation, each valid incoming state receives the upstream gradient scaled by the reciprocal of the number of valid incoming states at that node. The yellow arrows in Fig.~\ref{fig:paradigm}(b) illustrate this gradient flow.

\subsubsection{Inductive Kriging}

After training, we obtain the learned propagation rule $P_{\theta}^{*}$, which is independent of specific node identities and propagation depths. Therefore, it can be directly applied to focal-specific test subgraphs in the same manner as during training.

For each unobserved focal node $v_j\in\mathcal{V}_u$, we use the test subgraph $\mathcal{G}_{v_j}^{\mathrm{te}}$ extracted during adaptive subgraph extraction and denote its maximum propagation depth by $H_{v_j}$. Starting from the observed boundary nodes, the learned rule $P_{\theta}^{*}$ is repeatedly applied hop by hop toward the focal node. Propagated states from different branches are progressively aggregated until the propagated state of the focal node is obtained.

Finally, the propagated state at the focal node is used to reconstruct  the target traffic series. Since the same learned propagation rule is directly reused on previously unseen focal nodes and their corresponding test subgraphs without model retraining, the proposed paradigm enables inductive traffic kriging.

\begin{figure}[t]
    \centering
    \includegraphics[width=\linewidth]{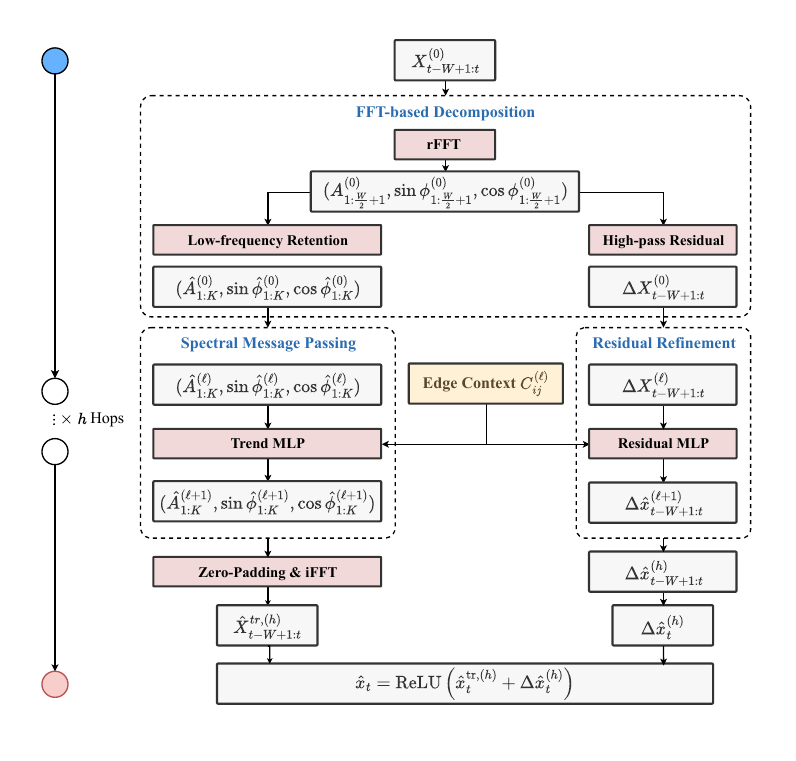}
    \caption{Overview of SPN.}
    \label{fig:model}
\end{figure}
  
\subsection{Spectral Propagation Network}\label{sec:model}

In the proposed paradigm, $P_{\theta}$ represents an abstract node-to-node propagation rule. In this section, we instantiate the paradigm with a Spectral Propagation Network (SPN), which consists of three components: FFT-based Spectral Decomposition, Spectral Message Passing, and Residual Refinement. For notational simplicity, in the following, $\mathbf{x}_i\in\mathbb{R}^{W}$ denotes a traffic window of length $W$ at node $v_i$.

\subsubsection{FFT-Based Spectral Decomposition}

As introduced in Section~\ref{sec:preliminary}, we transform the traffic window $\mathbf{x}_i$ into the frequency domain using the real-valued Fast Fourier Transform (rFFT): 
\begin{equation}
    \mathbf{z}_i=\operatorname{rFFT}(\mathbf{x}_i),
\end{equation}
where $\mathbf{z}_i$ denotes the non-redundant frequency-domain representation of the traffic window at node $v_i$.

To model the transferable traffic dynamics in the spectral domain, SPN retains the first $K$ low-frequency coefficients as the spectral features for subsequent propagation. These retained components primarily capture the slowly varying temporal patterns of the traffic series and therefore naturally form the spectral trend representation. Specifically, we construct a low-frequency spectrum as
\begin{equation}
    \tilde{z}_{i,k}=
    \begin{cases}
        z_{i,k}, & 0\leq k<K,\\
        0, & K\leq k<\lfloor W/2\rfloor+1,
    \end{cases}
\end{equation}
where $K$ denotes the number of retained low-frequency bins. By suppressing the higher-frequency components, $\tilde{\mathbf{z}}_i$ preserves the slowly varying temporal patterns, which are treated as the trend component in our model.

The corresponding trend signal in the time domain is reconstructed through the inverse rFFT:
\begin{equation}
    \tilde{\mathbf{x}}_i
    =
    \operatorname{irFFT}(\tilde{\mathbf{z}}_i).
\end{equation}
The remaining temporal variations are then represented by the residual:
\begin{equation}
    \Delta\mathbf{x}_i
    =
    \mathbf{x}_i-\tilde{\mathbf{x}}_i.
\end{equation}
The spectral trend representation $\tilde{\mathbf{z}}_i$ and the temporal residual $\Delta\mathbf{x}_i$ are subsequently modeled by the Spectral Message Passing and Residual Refinement modules, respectively.

\subsubsection{Spectral Message Passing}

After obtaining the retained low-frequency components, we perform message passing directly in the spectral domain. We first describe the spectral representation of the traffic window and then introduce how it is propagated.

\textbf{Spectrum Representation.}
Each retained Fourier coefficient is complex-valued and can be represented by its amplitude and phase. For the $k$-th retained frequency component, we have
\begin{equation}
    \tilde{z}_{i,k}
    =
    A_{i,k}e^{j\phi_{i,k}},
\end{equation}
where $A_{i,k}=|\tilde{z}_{i,k}|$ and $\phi_{i,k}=\arg(\tilde{z}_{i,k})$ denote its amplitude and phase, respectively. Directly using $\phi_{i,k}$ as a scalar feature introduces a discontinuity because phase is periodic, e.g., $-\pi$ and $\pi$ correspond to the same direction but are numerically far apart. To preserve this periodic structure and provide a continuous representation for the neural network, we encode each phase using its sine and cosine values. Accordingly, for an observed boundary node $v_i$, its initial spectral state is constructed from the retained low-frequency spectrum as
\begin{equation}
    \mathbf{s}_i^{(0)}
    =
    \left[
        \mathbf{A}_{i,0:K-1}^{(0)},
        \sin\boldsymbol{\phi}_{i,0:K-1}^{(0)},
        \cos\boldsymbol{\phi}_{i,0:K-1}^{(0)}
    \right].
\end{equation}
The pair $(\sin\phi,\cos\phi)$ uniquely specifies the phase on the unit circle while avoiding the boundary discontinuity of the raw phase angle.

\textbf{Spectral Propagation.}
Given a directed propagation edge $v_i\rightarrow v_j$, we concatenate the spectral state of the source node with the corresponding transition context:
\begin{equation}
    \mathbf{h}_{ij}^{(\ell)}
    =
    \left[
        \mathbf{s}_{i}^{(\ell)}
        \Vert
        \mathbf{c}_{ij}
    \right],
\end{equation}
where $\Vert$ denotes concatenation and $\mathbf{c}_{ij}$ denotes the transition context associated with the propagation from node $v_i$ to node $v_j$. The transition context incorporates the road metadata of the source and destination nodes, their pairwise road relations, local structural information, and temporal context. These contextual features are directly concatenated with the spectral state, allowing the propagation rule to condition the node-to-node state transition on both traffic dynamics and road-network characteristics. The specific features included in $\mathbf{c}_{ij}$ are detailed in Section~\ref{sec:expset}.

The concatenated representation is then fed into the shared Trend MLP:
\begin{equation}
    \hat{\mathbf{s}}_{ij}^{(\ell+1)}
    =
    P_{\theta}^{\mathrm{tr}}
    \left(
        \mathbf{h}_{ij}^{(\ell)}
    \right),
\end{equation}
where $P_{\theta}^{\mathrm{tr}}$ denotes the shared spectral propagation rule and $\hat{\mathbf{s}}_{ij}^{(\ell+1)}$ denotes the predicted amplitude--phase representation of the spectral message  from node $v_i$ to node $v_j$. Following the proposed propagation rule learning paradigm, the same $P_{\theta}^{\mathrm{tr}}$ is reused across different propagation edges, hops, and focal-specific subgraphs.

When multiple propagated spectral states arrive at the same destination node, they are aggregated to form its spectral state for the next propagation hop. However, directly averaging the amplitude--phase representation would treat the amplitude and phase components independently and would not preserve their joint complex spectral structure. Before aggregation, the predicted amplitude--phase representation $\hat{\mathbf{s}}_{ij}^{(\ell+1)}$ is therefore converted into a complex spectral message:

\begin{equation}
    \hat{\mathbf{z}}_{ij}^{(\ell+1)}
    =
    \mathcal{C}
    \left(
        \hat{\mathbf{s}}_{ij}^{(\ell+1)}
    \right),
\end{equation}

where $\mathcal{C}(\cdot)$ denotes the complex reconstruction operator, which constrains the amplitude to be non-negative, normalizes the sine and cosine phase components, and reconstructs the corresponding complex Fourier coefficients. For the DC component, the phase is fixed to zero so that the corresponding Fourier coefficient remains real-valued. We then perform mean aggregation directly in the complex spectral domain: 

\begin{equation}
    \mathbf{z}_{j}^{(\ell+1)}
    =
    \frac{1}{|\mathcal{N}_{j}^{(\ell)}|}
    \sum_{i\in\mathcal{N}_{j}^{(\ell)}}
    \hat{\mathbf{z}}_{ij}^{(\ell+1)},
\end{equation}
where $\mathcal{N}_{j}^{(\ell)}$ denotes the set of source nodes whose spectral states are propagated to node $v_j$ at hop $\ell$. The aggregated complex spectral state $\mathbf{z}_{j}^{(\ell+1)}$ is then converted back to its amplitude--phase representation $\mathbf{s}_{j}^{(\ell+1)}$ by extracting the amplitude and the normalized sine and cosine phase components. The resulting $\mathbf{s}_{j}^{(\ell+1)}$ serves as the source spectral representation for the next propagation hop.

During training, the complete multi-hop propagation process is performed before computing the prediction loss at the focal nodes. The resulting loss is then backpropagated through the entire propagation process. Since the same Trend MLP $P_{\theta}^{\mathrm{tr}}$ is repeatedly applied across different edges, hops, and focal-specific subgraphs, gradient contributions from all applications are accumulated to update the same set of parameters. Therefore, $P_{\theta}^{\mathrm{tr}}$ is optimized end-to-end according to the final focal-node predictions rather than independently at each propagation hop.

\subsubsection{Residual Refinement}

Although the retained low-frequency components capture the dominant temporal trend of traffic states, removing the higher-frequency components inevitably discards short-term temporal variations. To complement the spectral trend propagation, we introduce a Residual Refinement module to model the remaining temporal information directly in the time domain.

As defined in the FFT-based spectral decomposition, the temporal residual of node $v_i$ is obtained by subtracting the reconstructed low-frequency trend from the original traffic window:
\begin{equation}
    \Delta\mathbf{x}_i
    =
    \mathbf{x}_i-\tilde{\mathbf{x}}_i.
\end{equation}
Different from the trend branch, which propagates low-frequency spectral representations, the residual branch directly propagates the temporal residual without further frequency transformation.

Given a directed propagation edge $v_i\rightarrow v_j$, the residual state of the source node is concatenated with the same transition context used in spectral propagation:
\begin{equation}
    \mathbf{r}_{ij}^{(\ell)}
    =
    \left[
        \Delta\mathbf{x}_{i}^{(\ell)}
        \Vert
        \mathbf{c}_{ij}
    \right],
\end{equation}
where $\mathbf{c}_{ij}$ incorporates the road metadata, pairwise road relations, local structural information, and temporal context associated with the propagation from $v_i$ to $v_j$. The concatenated representation is fed into a shared Residual MLP:
\begin{equation}
    \Delta\hat{x}_{ij}^{(\ell+1)}
    =
    P_{\psi}^{\mathrm{res}}
    \left(
        \mathbf{r}_{ij}^{(\ell)}
    \right),
\end{equation}
where $P_{\psi}^{\mathrm{res}}$ denotes the residual propagation rule. Similar to the Trend MLP, the same Residual MLP is reused across different propagation edges, hops, and focal-specific subgraphs. When multiple residual messages arrive at the same destination node, they are aggregated following the same propagation structure as the spectral branch.

The residual branch is trained after the spectral trend propagation stage. Specifically, we first train the Trend MLP and select the best model according to the validation performance. Its parameters are then frozen, and only the Residual MLP is optimized in the refinement stage. Notably, the Residual MLP does not directly receive the reconstructed trend window as input. This prevents the residual branch from trivially recovering the original traffic signal from the combination of trend and residual, thereby preserving the spectral inductive bias introduced by the trend branch.

After $h$ propagation hops, the Residual MLP produces the residual correction $\Delta\hat{x}_{q}^{(h)}$ at the focal node $q$. The final traffic-state prediction is obtained by combining the frozen trend prediction with the residual correction:
\begin{equation}
    \hat{x}_{q}
    =
    \operatorname{ReLU}
    \left(
        \hat{x}_{q}^{\mathrm{tr},(h)}
        +
        \Delta\hat{x}_{q}^{(h)}
    \right).
\end{equation}

\section{Experiment}\label{sec:exp}

In this section, we first introduce the TraffiKrig benchmark in Section~\ref{sec:benchmark} and describe the experimental settings in Section~\ref{sec:expset}. We then evaluate LINK's kriging performance, zero-shot transferability, and efficiency in Sections~\ref{sec:performance}, \ref{sec:zeroshot}, and~\ref{sec:efficiency}, respectively. In Section~\ref{sec:metadata}, we further analyze the role of road metadata as contextual information for traffic state transitions. We then examine the contributions of the key designs in LINK through ablation studies in Section~\ref{sec:ablation}, followed by a sensitivity analysis of the main hyperparameters in Section~\ref{sec:sensitivity}. Finally, Section~\ref{sec:case} presents a case study to qualitatively illustrate the reconstruction behavior of LINK.

\subsection{TraffiKrig Dataset}\label{sec:benchmark}

To evaluate traffic kriging under explicit road-network constraints, we construct \textbf{TraffiKrig} from TraffiDent~\cite{gou2026traffident} by selecting sensors within the target districts and organizing them into four regional subsets, while retaining the corresponding road-network topology and road-specific sensor metadata. 


Rather than uniformly sampling sensors from the original network, we construct each TraffiKrig subset through deterministic spatial and structural filtering. In particular, we evaluate the two-dimensionality of each candidate region using the eigenvalue ratio of the covariance matrix of local sensor coordinates. Regions with extremely anisotropic or nearly collinear sensor distributions are filtered out to avoid selecting sensors concentrated along a single road corridor. We further require sufficient local sensor density and freeway diversity when determining seed regions. Following this procedure, we obtain four regional subsets, denoted as TraffiKrig-03/04/07/08, containing 381, 487, 1,151, and 512 nodes, respectively. To ensure temporal consistency across these subsets, all traffic observations are aligned to a common period from January 1 to April 30, 2023.

Compared with the commonly used PEMS dataset~\cite{guo2021astgnn}, TraffiKrig retains sensors deployed on heterogeneous road segments, including mainlines, ramps, and HOV lanes, together with road type, direction, and freeway information. Such heterogeneous node types are important for evaluating traffic kriging, because they allow us to examine whether a model can infer traffic states across road segments with different functional characteristics and traffic scales. The road network is represented as a directed weighted graph. For each node, the adjacency structure is sparsified by separately retaining the top-$K$ incoming and top-$K$ outgoing neighbors according to the edge weights, where we set $K_\mathcal{G}=3$ as the default configuration. By retaining both road-network topology and road-specific metadata, TraffiKrig enables the systematic evaluation of how traffic kriging methods exploit realistic road context.

\begin{table}[h]
\centering
\caption{Summary of Parameters}
\label{tab:parameters}
\begin{tabular}{ll}
\toprule
\textbf{Parameter Name} & \textbf{Description} \\
\midrule
Length of Time Window $W$ & 24 \\
Number of Training Rounds & 50 \\
Early Stopping Patience & 10 \\
Random Seed & 42/43/44\\
\midrule

Retained Frequency Bins $K$ & 4 (LINK)\\
Hidden Dimension & 64 (LINK) \\

\midrule
Learning Rate
& 0.003 (Most) \\
& 0.01 (GRIN, INCREASE)\\
Batch Size
& 4 (KITS) \\
& 64 (others) \\
\bottomrule
\end{tabular}
\end{table}

\subsection{Experiment Settings}\label{sec:expset}

For all datasets, we randomly split the sensor nodes into training, validation, and test sets with a ratio of 7:1:2, respectively. The node split is kept fixed across three random seeds, and we report the mean and standard deviation of the evaluation metrics. The traffic observations of validation and test sensors are held out, while their road-network topology and road-specific metadata remain available. We use the validation set for model selection and report the performance on the test sensors. 
We use three types of static road metadata: road type, travel direction, and freeway identity. Road type and travel direction are encoded categorically for both the source and destination nodes, while freeway identity is used only to determine whether a propagation edge connects nodes on the same freeway rather than being directly encoded. We use Smooth L1 loss for training. The Trend MLP is supervised on the decoded low-frequency trend, while the Residual MLP is optimized using both the residual prediction loss and the final reconstruction loss, with the latter weighted by $\lambda=0.5$.

\textbf{Baselines.}
We compare LINK with a broad set of graph-based and spatio-temporal kriging baselines, including GCN~\cite{kipf2016gcn}, Okriging, GIN~\cite{xu2018gin}, IGNNK~\cite{wu2021IGNNK}, SATCN~\cite{wu2021satcn}, INCREASE~\cite{zheng2023increase}, DualSTN~\cite{hu2023dualstn}, IAGCN~\cite{wei2024iagcn}, GRIN~\cite{cini2021grin}, STA-GANN~\cite{li2025stagann}, and KITS~\cite{xu2025kits}. Except for KITS, we follow the implementations provided in STA-GANN~\cite{li2025stagann}. For KITS, we adopt the official implementation from its authors' repository.

\textbf{Hyperparameters.}
The main experimental hyperparameters are summarized in Table~\ref{tab:parameters}. We use a historical window of $W=24$ time steps for all methods. For SPN, the lowest $K=4$ frequency bins are retained for spectral propagation. For SPN, each optimization step jointly processes seven focal-specific propagation graphs centered at different pseudo-focal nodes. During each training epoch, all training focal nodes are traversed and grouped into such sets of seven. Both rFFT and irFFT are implemented with orthonormal normalization.

\textbf{Metrics.}
We evaluate kriging performance using Mean Absolute Error (MAE), Root Mean Squared Error (RMSE), and Mean Absolute Percentage Error (MAPE), where lower values indicate better performance.

\begin{table*}[t]
\centering
\caption{Overall performance comparison on TraffiKrig.}
\label{tab:overall_results}
\renewcommand{\arraystretch}{1.15}
\setlength{\tabcolsep}{3.0pt}
\resizebox{\textwidth}{!}{%
\begin{tabular}{l|ccc|ccc|ccc|ccc}

\hline
\textbf{Model} & \multicolumn{3}{c|}{\textbf{03}} & \multicolumn{3}{c|}{\textbf{04}} & \multicolumn{3}{c|}{\textbf{07}} & \multicolumn{3}{c}{\textbf{08}} \\
\cline{2-13}
& \textbf{MAE} & \textbf{RMSE} & \textbf{MAPE} & \textbf{MAE} & \textbf{RMSE} & \textbf{MAPE} & \textbf{MAE} & \textbf{RMSE} & \textbf{MAPE} & \textbf{MAE} & \textbf{RMSE} & \textbf{MAPE} \\

\hline
GCN & $93.40 \pm 1.22$ & $123.56 \pm 1.11$ & $372.65 \pm 18.78$ & $115.99 \pm 2.26$ & $155.76 \pm 0.98$ & $510.56 \pm 13.55$ & $140.88 \pm 1.30$ & $178.26 \pm 1.31$ & $473.07 \pm 2.39$ & $94.40 \pm 0.88$ & \underline{119.17} $\pm$ 1.01 & $415.17 \pm 2.11$ \\
GIN & $93.63 \pm 0.68$ & $128.86 \pm 0.99$ & $380.26 \pm 3.34$ & $105.66 \pm 0.26$ & $134.39 \pm 1.28$ & $449.53 \pm 20.73$ & $129.25 \pm 0.94$ & $171.13 \pm 0.41$ & $455.97 \pm 24.23$ & $95.37 \pm 1.10$ & $127.32 \pm 2.35$ & $357.13 \pm 2.75$ \\
Okriging & $93.22 \pm 0.17$ & $131.52 \pm 0.36$ & $288.00 \pm 10.31$ & $118.51 \pm 0.14$ & $151.14 \pm 0.11$ & $388.82 \pm 0.42$ & $139.97 \pm 0.02$ & $184.07 \pm 0.19$ & $446.32 \pm 2.91$ & $109.67 \pm 0.36$ & $141.42 \pm 0.41$ & $386.13 \pm 3.86$ \\
GRIN & $99.54 \pm 1.30$ & $139.16 \pm 3.07$ & $304.03 \pm 9.19$ & $117.73 \pm 5.08$ & $159.47 \pm 5.23$ & $377.82 \pm 32.20$ & $127.30 \pm 2.35$ & $184.98 \pm 2.05$ & $299.70 \pm 13.39$ & $96.46 \pm 2.47$ & $136.02 \pm 3.45$ & $311.80 \pm 18.81$ \\
SATCN & $105.59 \pm 0.88$ & $154.16 \pm 1.89$ & $280.00 \pm 6.88$ & $117.16 \pm 0.10$ & $152.63 \pm 0.25$ & $400.99 \pm 6.44$ & $151.10 \pm 0.11$ & $201.34 \pm 1.22$ & $422.41 \pm 3.24$ & $101.46 \pm 0.31$ & $134.96 \pm 0.90$ & $353.68 \pm 10.51$ \\
INCREASE & $100.83 \pm 2.87$ & $147.80 \pm 2.61$ & $217.20 \pm 17.63$ & $106.57 \pm 5.82$ & $148.55 \pm 7.07$ & $394.08 \pm 59.19$ & $134.01 \pm 1.07$ & $193.11 \pm 2.18$ & $310.69 \pm 17.23$ & $119.05 \pm 5.33$ & $159.55 \pm 11.57$ & $365.52 \pm 15.88$ \\
DualSTN & $112.86 \pm 9.03$ & $162.17 \pm 12.81$ & $274.37 \pm 27.62$ & $177.89 \pm 63.27$ & $255.77 \pm 111.19$ & $518.36 \pm 183.40$ & $175.08 \pm 10.11$ & $242.48 \pm 11.46$ & $387.22 \pm 32.28$ & $145.31 \pm 18.65$ & $197.27 \pm 22.99$ & $361.45 \pm 146.23$ \\
IAGCN & $100.88 \pm 2.16$ & $148.00 \pm 2.23$ & \underline{209.15} $\pm$ 16.36 & $119.84 \pm 6.54$ & $158.20 \pm 9.22$ & $357.42 \pm 9.58$ & $129.91 \pm 0.47$ & $185.32 \pm 5.76$ & $279.80 \pm 54.90$ & $97.55 \pm 1.69$ & $140.73 \pm 2.67$ & \underline{203.90} $\pm$ 20.04 \\
IGNNK & $113.36 \pm 2.01$ & $162.97 \pm 3.04$ & $330.76 \pm 18.70$ & $119.05 \pm 1.79$ & $165.97 \pm 3.64$ & $385.18 \pm 56.80$ & $125.54 \pm 1.74$ & $175.11 \pm 3.46$ & $288.87 \pm 20.30$ & $101.45 \pm 2.83$ & $137.73 \pm 2.92$ & $306.09 \pm 18.72$ \\
STA-GANN & \underline{86.84} $\pm$ 2.23 & \underline{120.86} $\pm$ 4.09 & $235.05 \pm 25.04$ & \underline{101.07} $\pm$ 3.13 & \underline{133.04} $\pm$ 2.67 & \underline{355.72} $\pm$ 26.07 & \underline{119.28} $\pm$ 1.36 & \underline{166.48} $\pm$ 1.06 & \underline{277.23} $\pm$ 27.88 & \underline{88.09} $\pm$ 0.99 & $120.26 \pm 2.66$ & $274.59 \pm 17.29$ \\
KITS & $92.94 \pm 0.66$ & $131.36 \pm 0.69$ & $229.95 \pm 35.03$ & $101.85 \pm 1.96$ & $134.72 \pm 10.48$ & $456.93 \pm 53.20$ & $132.76 \pm 0.96$ & $191.57 \pm 1.77$ & $286.04 \pm 10.81$ & $111.25 \pm 2.27$ & $146.51 \pm 0.95$ & $349.10 \pm 14.47$ \\
\hline\hline
\textbf{LINK} & \textbf{39.26} $\pm$ 1.10 & \textbf{63.09} $\pm$ 2.04 & \textbf{67.95} $\pm$ 4.46 & \textbf{44.94} $\pm$ 0.99 & \textbf{66.71} $\pm$ 1.35 & \textbf{74.86} $\pm$ 4.52 & \textbf{58.89} $\pm$ 0.52 & \textbf{102.94} $\pm$ 1.49 & \textbf{65.44} $\pm$ 3.91 & \textbf{37.16} $\pm$ 0.16 & \textbf{54.74} $\pm$ 0.35 & \textbf{87.16} $\pm$ 5.75 \\
\hline
\end{tabular}%
}
\end{table*}

\subsection{Overall Performance}\label{sec:performance}

Table~\ref{tab:overall_results} reports the performance of all compared baselines under their original settings and LINK with road metadata on the four TraffiKrig subsets.

LINK consistently achieves the best performance across all four subsets and all three evaluation metrics. Compared with the strongest baseline on each subset, LINK achieves average reductions of 54.7\%, 47.5\%, and 70.0\% in MAE, RMSE, and MAPE, respectively, across the four TraffiKrig subsets. These substantial improvements indicate that the proposed propagation rule learning paradigm can more accurately recover traffic magnitudes at unobserved locations across heterogeneous road-network regions, as further illustrated by the case study in Section~\ref{sec:case}.

The strongest baseline varies across different subsets and evaluation metrics, indicating that existing methods do not consistently dominate under different road-network conditions. In contrast, LINK achieves the lowest MAE, RMSE, and MAPE on every TraffiKrig subset, demonstrating more consistent reconstruction performance across different regions and traffic scales.

The particularly large improvement in MAPE indicates that LINK substantially reduces relative reconstruction errors. Since MAPE places greater emphasis on errors at low-magnitude observations, this result also suggests improved reconstruction of low-volume traffic states, which is important when traffic magnitudes vary substantially across heterogeneous road segments.

\begin{table}[h]
\centering
\caption{Zero-shot transfer performance of LINK.}
\label{tab:zero_shot}

\renewcommand{\arraystretch}{1.15}
\setlength{\tabcolsep}{12pt}

\begin{tabular}{l|ccc}
\hline
\textbf{Dataset}
& \textbf{MAE}
& \textbf{RMSE}
& \textbf{MAPE} \\
\hline
TraffiKrig-04 & 50.17 & 71.98 & 70.91 \\
TraffiKrig-07 & 72.78 & 116.74 & 62.65 \\
TraffiKrig-08 & 45.23 & 69.27 & 79.75 \\
\hline
\end{tabular}
\end{table}

\subsection{Zero-shot Experiment}\label{sec:zeroshot}

To further investigate whether LINK learns transferable propagation rules rather than region-specific spatial correlations, we conduct a zero-shot experiment. Specifically, LINK is trained exclusively on TraffiKrig-03 and directly evaluated on the other three subsets without any target-region training or fine-tuning. Table~\ref{tab:zero_shot} reports the corresponding zero-shot transfer performance.

Unlike existing GNN-based kriging baselines that rely on region-specific graph structures and spatial dependencies, LINK learns a shared propagation rule that enables direct generalization to previously unseen road networks. Remarkably, even under the zero-shot setting, LINK consistently outperforms the strongest baselines evaluated under their original in-domain settings. Specifically, LINK achieves MAE reductions of 50.36\%, 38.98\%, and 48.65\% on other three subsets, respectively. These results provide further evidence that LINK captures transferable propagation patterns and can generalize across unseen road networks without parameter updates.

\begin{table}[h]
\centering
\small
\caption{Efficiency comparison on TraffiKrig-03.}
\label{tab:efficiency_comparison}

\renewcommand{\arraystretch}{1.15}
\setlength{\tabcolsep}{5pt}

\begin{tabular}{l|c|c|c}
\hline
\textbf{Model}
& \textbf{Training (s)}
& \textbf{Inference (ms)}
& \textbf{Peak GPU (MiB)} \\
\hline
DualSTN  & 2030.82  & 1.605 & 17,248.89 \\
GCN      & 637.65   & 0.719 & 337.07 \\
GIN      & 1253.18  & 0.717 & 355.63 \\
GRIN     & 2931.06  & 1.352 & 1,973.46 \\
IAGCN    & 6582.55  & 9.146 & 11,484.62 \\
IGNNK    & 560.80   & 0.630 & 356.51 \\
INCREASE & 17,226.74 & 4.728 & 937.07 \\
Okriging & 518.87   & 0.625 & 293.68 \\
SATCN    & 451.58   & 0.628 & 556.15 \\
STA-GANN  & 564.72   & 0.891 & 2,587.50 \\
KITS     & 607.74   & 1.870 & 1,055.07 \\
\hline
\textbf{LINK} 
& \textbf{151.30} 
& \textbf{0.084} 
& \textbf{190.11} \\
\hline
\end{tabular}
\end{table}

\begin{figure}[t]
    \centering
        \includegraphics[width=\linewidth]{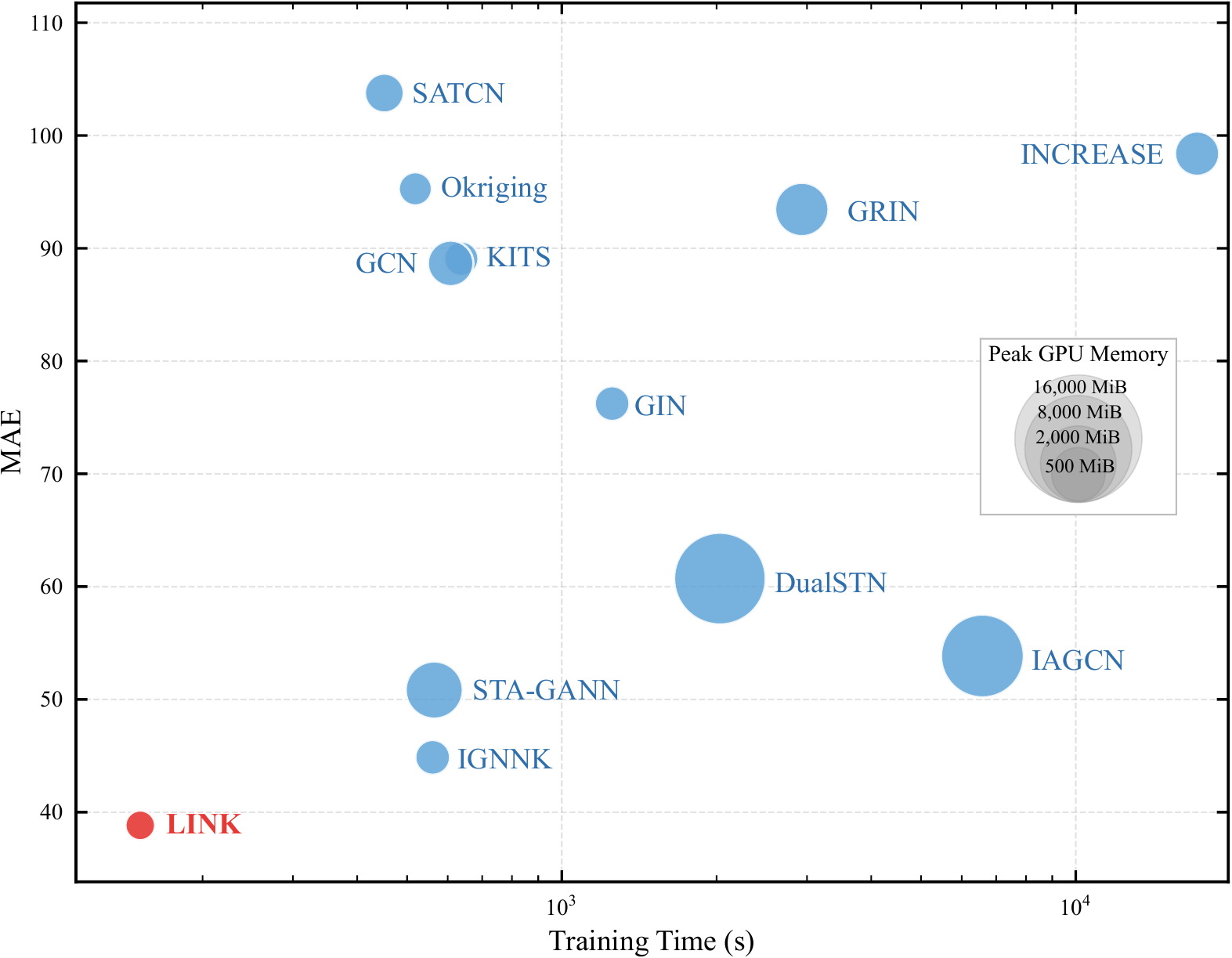}
        \caption{Efficiency-performance trade-off on TraffiKrig-03.}
        \label{fig:efficiency}
\end{figure}

\subsection{Efficiency Analysis}\label{sec:efficiency}

We further evaluate the computational efficiency of LINK and baseline methods on TraffiKrig-03, and analyze the trade-off between reconstruction accuracy and computational cost. All methods are executed on the same NVIDIA RTX 4090 GPU with Python 3.10, using the hyperparameter settings listed in Table~\ref{tab:parameters}. 

For each method, we report the total training time, inference time per sample, and peak GPU memory usage during training. The total training time is measured as the accumulated runtime over all training epochs required by the method. Peak GPU memory is measured as the maximum GPU memory allocated during training. The total training time of LINK is computed as the sum of the training time of the Trend MLP and Residual MLP. Although Okriging is inherently non-parametric, the adopted STA-GANN implementation evaluates it using the same epoch-based training pipeline as the learning-based baselines.

As shown in Table~\ref{tab:efficiency_comparison}, LINK exhibits substantially lower computational overhead than all compared baselines. Its total training time is 151.30 seconds, which is only 33.5\% of SATCN, the fastest baseline, corresponding to an approximately 3.0$\times$ speedup. LINK also requires the lowest peak GPU memory, using 190.11 MiB compared with 293.68 MiB for the most memory-efficient baseline. In addition, LINK achieves an inference time of only 0.084 ms per sample, which is approximately 7.4$\times$ faster than the fastest baseline, Okriging. This efficiency is mainly attributed to the design of the LINK paradigm, where a single lightweight SPN is repeatedly reused to model node-to-node traffic state transitions across different propagation edges and depths. Moreover, focal-specific adaptive subgraphs restrict computation to the propagation structures relevant to each target node, avoiding unnecessary computation over the complete road network.

Fig.~\ref{fig:efficiency} further illustrates the trade-off between computational efficiency and kriging performance under the metadata-enhanced setting. Blue points denote metadata-enhanced baseline methods, while the red point represents LINK. Compared with metadata-enhanced baselines on TraffiKrig-03, LINK maintains the lowest MAE while requiring substantially less training time and GPU memory. This demonstrates that the proposed propagation rule learning paradigm can achieve an effective balance between reconstruction accuracy and computational efficiency. Such efficiency benefits from two key designs of LINK: the lightweight SPN, which learns a shared node-to-node transition rule across different propagation edges and depths, and the adaptive subgraph extraction mechanism, which avoids unnecessary propagation over regions outside the focal-specific subgraphs.

\begin{table*}[t]
\centering
\caption{Performance of metadata-enhanced baseline methods on TraffiKrig.}
\label{tab:baseline_with_metadata}
\renewcommand{\arraystretch}{1.15}
\setlength{\tabcolsep}{3.0pt}
\resizebox{\textwidth}{!}{%
\begin{tabular}{l|c|ccc|ccc|ccc|ccc}
\hline
\textbf{Model} & \textbf{Imp. (\%)} & \multicolumn{3}{c|}{\textbf{03}} & \multicolumn{3}{c|}{\textbf{04}} & \multicolumn{3}{c|}{\textbf{07}} & \multicolumn{3}{c}{\textbf{08}} \\
\cline{3-14}
& & \textbf{MAE} & \textbf{RMSE} & \textbf{MAPE} & \textbf{MAE} & \textbf{RMSE} & \textbf{MAPE} & \textbf{MAE} & \textbf{RMSE} & \textbf{MAPE} & \textbf{MAE} & \textbf{RMSE} & \textbf{MAPE} \\
\hline
GCN & 10.35& $88.20 \pm 2.09$ & $122.45 \pm 2.72$ & $341.58 \pm 10.57$ & $95.96 \pm 0.78$ & $125.28 \pm 0.94$ & $429.22 \pm 6.03$ & $129.55 \pm 1.52$ & $169.23 \pm 3.35$ & $395.54 \pm 2.73$ & $84.48 \pm 0.69$ & $107.53 \pm 1.76$ & $346.08 \pm 5.20$ \\
GIN & 24.07 & $76.35 \pm 0.88$ & $109.61 \pm 1.44$ & $283.22 \pm 13.72$ & $82.07 \pm 4.93$ & $114.80 \pm 5.97$ & $262.41 \pm 4.96$ & $96.06 \pm 1.55$ & $137.91 \pm 0.57$ & $289.94 \pm 14.30$ & $66.93 \pm 1.53$ & $95.58 \pm 0.57$ & $264.81 \pm 12.93$ \\
Okriging & -0.72 & $94.49 \pm 1.46$ & $137.98 \pm 1.59$ & $324.29 \pm 11.69$ & $119.47 \pm 0.57$ & $150.40 \pm 0.70$ & $389.67 \pm 15.92$ & $142.46 \pm 1.47$ & $193.80 \pm 0.74$ & $410.91 \pm 13.22$ & $108.51 \pm 2.04$ & $140.85 \pm 1.07$ & $403.69 \pm 23.52$ \\
GRIN & 1.52& $97.73 \pm 3.80$ & $137.91 \pm 4.55$ & $269.25 \pm 31.96$ & $117.01 \pm 4.15$ & $157.10 \pm 4.29$ & $318.19 \pm 34.04$ & $122.29 \pm 5.92$ & $179.88 \pm 6.32$ & $280.54 \pm 11.20$ & $96.73 \pm 1.73$ & $139.20 \pm 1.19$ & $314.94 \pm 15.49$ \\
SATCN & 0.85 & $104.27 \pm 0.66$ & $153.78 \pm 2.41$ & $257.92 \pm 23.10$ & $119.32 \pm 0.50$ & $155.65 \pm 0.36$ & $393.35 \pm 8.73$ & $143.59 \pm 0.72$ & $202.73 \pm 1.22$ & $333.74 \pm 11.02$ & $102.46 \pm 0.71$ & $143.12 \pm 1.07$ & $345.53 \pm 8.27$ \\
INCREASE & -0.69& $99.99 \pm 7.60$ & $146.04 \pm 8.86$ & $251.18 \pm 14.89$ & $110.42 \pm 0.71$ & $151.40 \pm 4.12$ & $336.45 \pm 10.66$ & $132.26 \pm 0.93$ & $193.71 \pm 2.57$ & $260.72 \pm 52.57$ & $120.60 \pm 10.78$ & $161.42 \pm 13.27$ & $390.57 \pm 58.34$ \\
DualSTN & 53.33& $62.92 \pm 7.78$ & $97.76 \pm 9.79$ & $107.35 \pm 15.57$ & $77.86 \pm 17.20$ & $108.81 \pm 21.06$ & $154.33 \pm 116.92$ & $94.47 \pm 2.98$ & $144.43 \pm 1.57$ & $166.18 \pm 6.77$ & $48.25 \pm 7.61$ & $72.41 \pm 10.29$ & $107.55 \pm 25.92$ \\
IAGCN & 50.28& $54.39 \pm 0.67$ & $84.97 \pm 0.55$ & \underline{85.87} $\pm$ 3.39 & $52.35 \pm 1.64$ & $75.97 \pm 2.66$ & \underline{91.30} $\pm$ 17.81 & $71.60 \pm 1.70$ & $118.55 \pm 1.07$ & \underline{83.11} $\pm$ 4.51 & $45.04 \pm 0.64$ & $65.30 \pm 0.68$ & \underline{97.11} $\pm$ 4.58 \\
IGNNK & 57.02& \underline{44.92} $\pm$ 1.71 & \underline{69.48} $\pm$ 1.86 & $101.68 \pm 2.61$ & \underline{51.18} $\pm$ 1.92 & \underline{72.71} $\pm$ 2.36 & $94.73 \pm 14.90$ & \underline{62.54} $\pm$ 1.43 & \underline{110.97} $\pm$ 2.98 & $101.00 \pm 4.48$ & \underline{40.05} $\pm$ 2.48 & \underline{56.53} $\pm$ 2.91 & $110.90 \pm 9.17$ \\
STA-GANN & 46.38& $49.21 \pm 2.25$ & $75.18 \pm 4.96$ & $108.04 \pm 10.97$ & $53.82 \pm 5.80$ & $74.98 \pm 8.14$ & $118.62 \pm 23.28$ & $68.66 \pm 2.87$ & $111.03 \pm 3.22$ & $141.57 \pm 13.87$ & $41.39 \pm 3.00$ & $59.29 \pm 3.31$ & $121.76 \pm 14.15$ \\
KITS & -1.71& $92.74 \pm 6.99$ & $130.96 \pm 8.09$ & $234.59 \pm 22.87$ & $105.09 \pm 0.84$ & $144.62 \pm 2.24$ & $418.77 \pm 30.32$ & $138.12 \pm 2.27$ & $195.77 \pm 5.08$ & $320.43 \pm 27.57$ & $111.07 \pm 5.18$ & $149.03 \pm 7.15$ & $357.81 \pm 24.38$ \\
\hline\hline
\textbf{LINK} & -- & \textbf{39.26} $\pm$ 1.10 & \textbf{63.09} $\pm$ 2.04 & \textbf{67.95} $\pm$ 4.46 & \textbf{44.94} $\pm$ 0.99 & \textbf{66.71} $\pm$ 1.35 & \textbf{74.86} $\pm$ 4.52 & \textbf{58.89} $\pm$ 0.52 & \textbf{102.94} $\pm$ 1.49 & \textbf{65.44} $\pm$ 3.91 & \textbf{37.16} $\pm$ 0.16 & \textbf{54.74} $\pm$ 0.35 & \textbf{87.16} $\pm$ 5.75 \\
\hline
\end{tabular}%
}
\end{table*}

\subsection{Road Metadata Analysis on TraffiKrig}\label{sec:metadata}

As shown in Table~\ref{tab:baseline_with_metadata}, we further establish a metadata-enhanced evaluation setting on TraffiKrig by incorporating the provided road metadata into all baseline methods. Specifically, all baseline methods are extended with road metadata using the same feature-concatenation strategy adopted in SPN. This setting enables us to investigate whether road metadata provides additional information for traffic kriging beyond traffic observations and road-network topology. The table reports the performance of metadata-enhanced baselines and LINK under the same setting. The ``Imp.'' column denotes the average MAE improvement obtained by each baseline after incorporating road metadata into its original formulation.

Most baseline methods benefit from incorporating road metadata, demonstrating that road-specific information provides valuable auxiliary signals for traffic state reconstruction. In particular, a large proportion of baseline methods achieve over 40\% average MAE reduction, with the maximum improvement reaching 57.02\%. These results verify that road metadata can effectively complement historical traffic observations and network topology by providing additional semantic and structural information for the kriging task.

However, the effectiveness of metadata utilization varies substantially across different methods. Several baselines obtain limited improvements, while Okriging and INCREASE even show slight performance degradation after incorporating metadata. This difference indicates that the capability of effectively integrating external road information is highly dependent on the model architecture. For metadata-enhanced Okriging, we prepend a trainable metadata-fusion MLP before the original non-parametric interpolation core. The slight performance degradation suggests that this input-level fusion strategy does not necessarily improve reconstruction under a fixed interpolation rule. Meanwhile, some kriging-oriented methods already rely on specialized spatial or auxiliary-information modeling mechanisms, which may limit the additional benefits obtained from simple metadata concatenation.

Compared with all metadata-enhanced baselines, LINK achieves the lowest MAE and MAPE across all four TraffiKrig subsets. Specifically, LINK achieves average improvements of 9.5\% in MAE and 17.6\% in MAPE compared with the strongest metadata-enhanced baseline on each subset. These results demonstrate that the performance gains of LINK do not solely come from utilizing road metadata, but also benefit from the proposed propagation rule learning paradigm, which explicitly models transferable node-to-node traffic state transitions.

\begin{figure*}[t]
    \centering

    \subfloat[Maximum propagation hop.]{
        \includegraphics[width=0.31\textwidth]{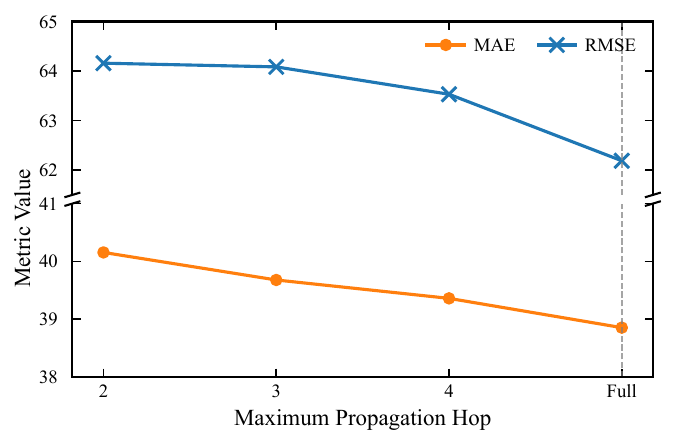}
    }
    \hfill
    \subfloat[Retained frequency bins $K$.]{
        \includegraphics[width=0.31\textwidth]{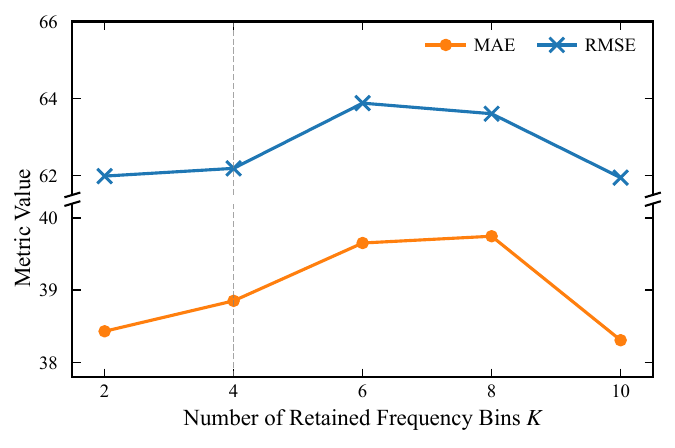}
    }
    \hfill
    \subfloat[Temporal window size $W$.]{
        \includegraphics[width=0.31\textwidth]{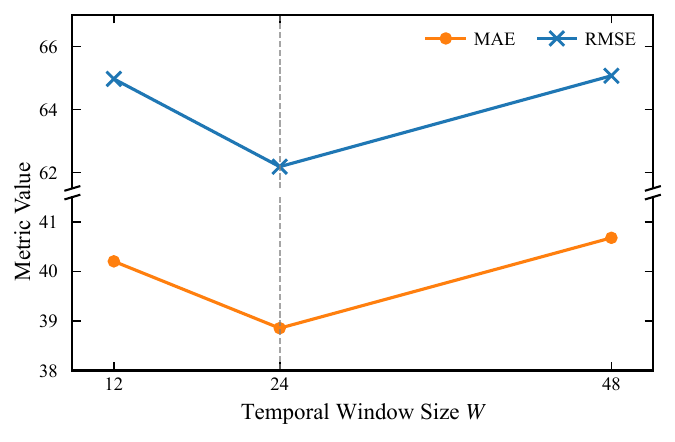}
    }

    \caption{Sensitivity analysis of spectral hyperparameters on TraffiKrig-03.}
    \label{fig:spectral_sensitivity}
\end{figure*}


\begin{figure*}[t]
    \centering

    \subfloat[Hidden dimension.\label{fig:hidden_sensitivity}]{
        \includegraphics[width=0.31\textwidth]{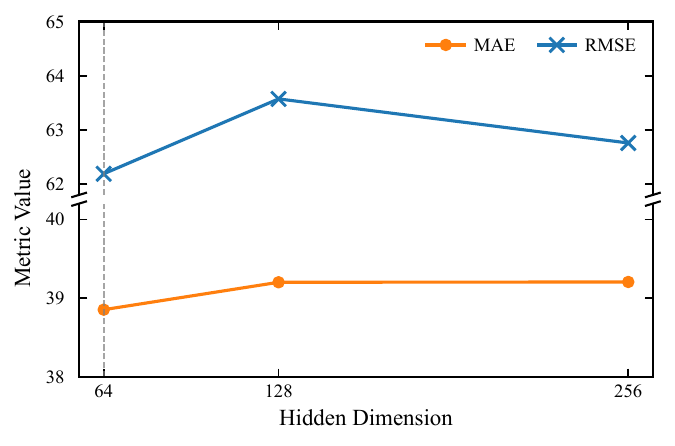}
    }
    \hfill
    \subfloat[Dropout rate.\label{fig:dropout_sensitivity}]{
        \includegraphics[width=0.31\textwidth]{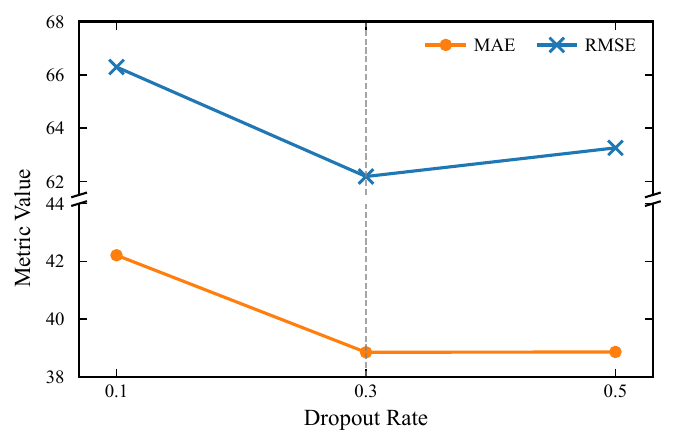}
    }
    \hfill
    \subfloat[Learning rate.\label{fig:lr_sensitivity}]{
        \includegraphics[width=0.31\textwidth]{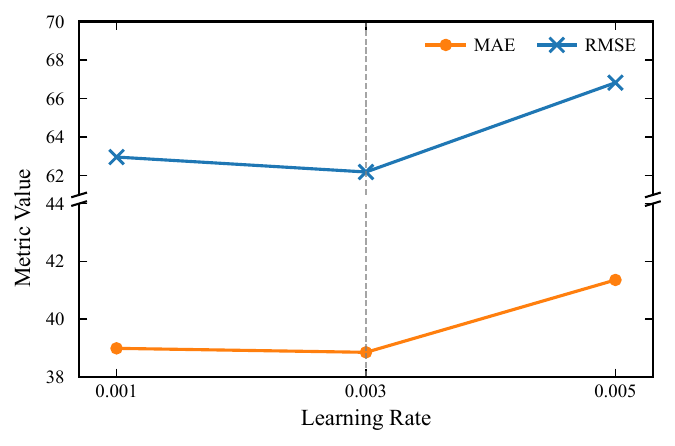}
    }

    \caption{Sensitivity analysis of model hyperparameters on TraffiKrig-03.}
    \label{fig:model_sensitivity}
    
\end{figure*}

\subsection{Ablation Study}\label{sec:ablation}


To investigate the contribution of the key designs in LINK, we conduct ablation studies on the TraffiKrig benchmark. Specifically, we consider three variants: \textit{Spectral GNN}, which replaces the focal-specific shared propagation mechanism with a two-layer GNN operating on the full graph while retaining the spectral representation; \textit{w/o Spectral Propagation}, which propagates the reconstructed trend in the time domain instead of directly propagating spectral states; and \textit{w/o Residual Refinement}, which removes the residual refinement stage. Since focal-specific subgraph construction and shared propagation rule learning are jointly designed as the core propagation paradigm of LINK, we use the Spectral GNN variant as a unified evaluation of their contribution.

As shown in Table~\ref{tab:ablation}, replacing the focal-specific shared propagation paradigm with a full-graph spectral GNN and removing the residual refinement stage both lead to consistent performance degradation, demonstrating the necessity of the proposed propagation architecture and residual correction mechanism. Notably, removing spectral propagation results in a substantially larger increase in MAPE than in RMSE. Although the RMSE slightly decreases on some subsets, the significant degradation in MAPE indicates that spectral-domain propagation is more effective in preserving the correct traffic magnitude, especially for low-volume observations where relative errors are more sensitive to scale mismatch. Overall, these results verify that all three components are important for the effectiveness of LINK, jointly contributing to accurate and robust traffic state reconstruction.

\begin{table}[h]
\centering
\caption{Ablation study on TraffiKrig.}
\label{tab:ablation}

\renewcommand{\arraystretch}{1.12}
\setlength{\tabcolsep}{5.5pt}

\begin{tabular}{llccc}
\toprule
\textbf{Subset}& \textbf{Variant}
& \textbf{MAE}
& \textbf{RMSE}
& \textbf{MAPE} \\
\midrule

\multirow{4}{*}{03}
& Full          & \textbf{38.83} & \textbf{62.66} & \textbf{64.57} \\
& Spectral GNN& 43.42 & 68.64 & 88.02 \\
& w/o Spectral Propagation& 40.78 & 64.57 & 85.36 \\
& w/o Residual Refinement& 41.30 & 65.37 & 85.41 \\
\midrule

\multirow{4}{*}{04}
& Full          & \textbf{45.71} & 67.75 & \textbf{76.10} \\
& Spectral GNN& 49.45 & 74.60 & 78.08 \\
& w/o Spectral Propagation& 46.14 & \textbf{67.29} & 87.15 \\
& w/o Residual Refinement& 50.00 & 73.32 & 78.41 \\
\midrule

\multirow{4}{*}{07}
& Full          & \textbf{57.74} & \textbf{101.16} & \textbf{65.89} \\
& Spectral GNN& 59.54 & 104.66 & 78.97 \\
& w/o Spectral Propagation& 60.37 & 104.97 & 82.31 \\
& w/o Residual Refinement& 61.58 & 104.70 & 70.97 \\
\midrule

\multirow{4}{*}{08}
& Full          & \textbf{38.42} & 56.65 & \textbf{79.71} \\
& Spectral GNN& 42.19 & 61.57 & 108.89 \\
& w/o Spectral Propagation& 38.99 & \textbf{56.13} & 104.38 \\
& w/o Residual Refinement& 42.08 & 62.24 & 84.90 \\
\bottomrule
\end{tabular}
\end{table}

\subsection{Hyperparameter Sensitivity Analysis}\label{sec:sensitivity}

We further investigate the sensitivity of LINK to several key hyperparameters on TraffiKrig-03 by varying one hyperparameter at a time while keeping the remaining settings fixed. As shown in Fig.~\ref{fig:spectral_sensitivity} and Fig.~\ref{fig:model_sensitivity}, each point corresponds to a specific hyperparameter configuration, while the orange and blue curves report MAE and RMSE, respectively, where lower values indicate better performance. The dashed vertical lines indicate the default configurations used in our experiments.

\subsubsection{Spectral Propagation Hyperparameters}

We first analyze the effect of propagation depth, which directly reflects one of the key differences between LINK and conventional fixed-depth GNNs. As shown in Fig.~\ref{fig:spectral_sensitivity}(a), both MAE and RMSE consistently decrease as the maximum propagation depth increases, while the full adaptive propagation setting achieves the best performance. This result indicates that informative observations may lie at different graph distances from different focal nodes, and restricting all nodes to a fixed shallow propagation depth can prevent useful information from reaching the target. In contrast, LINK operates on focal-specific subgraphs with varying propagation depths and repeatedly applies the same shared propagation rule across these depths, enabling the model to adapt its receptive field to each focal node.

We further examine the spectral representation controlled by the number of retained frequency bins $K$ and the temporal window size $W$. LINK remains relatively stable across different choices of $K$, indicating that the model is not highly sensitive to a particular spectral cutoff. This robustness is consistent with the dual-domain design of SPN, where the retained low-frequency components capture the slowly varying traffic trend, while information excluded from the spectral representation can be further recovered by the residual refinement stage. Meanwhile, $W$ determines the temporal scale represented by these retained frequency components. With a fixed $K$, increasing $W$ shifts the retained spectrum toward lower frequencies, whereas an overly short window provides insufficient temporal context for extracting meaningful low-frequency dynamics. The best performance at $W=24$ therefore reflects a balance between preserving sufficient temporal context and constructing an informative yet compact spectral representation.

\subsubsection{Model Hyperparameters}
We further evaluate the sensitivity of LINK to several model and optimization hyperparameters, including the hidden dimension, dropout rate, and learning rate, as shown in Fig.~\ref{fig:model_sensitivity}. Overall, LINK remains relatively stable across the evaluated configurations, indicating that its performance is not highly sensitive to these hyperparameters. Increasing the hidden dimension does not consistently improve performance, while moderate dropout and learning-rate settings yield stable results. We therefore adopt a hidden dimension of $64$, a dropout rate of $0.3$, and a learning rate of $0.003$ as the default configuration.

\subsection{Case Study}\label{sec:case}

\begin{figure}[h]
    \centering
    \includegraphics[width=\linewidth]{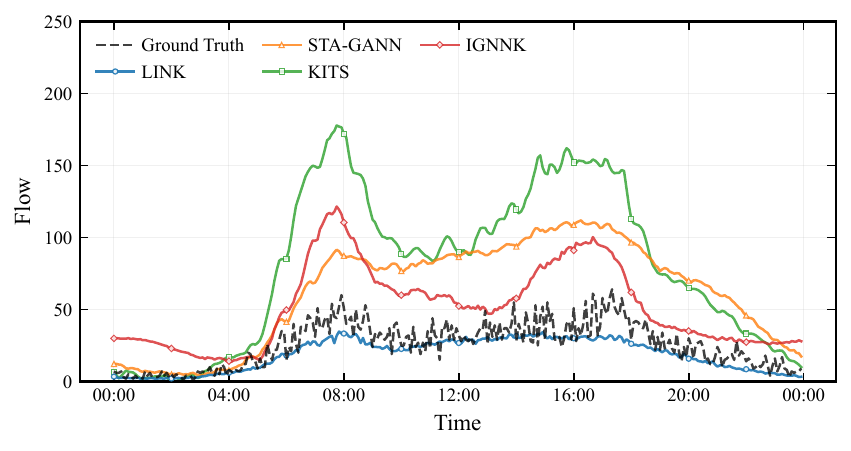}
    \caption{Kriging result with original baselines}
    \label{fig:casewithout}
\end{figure}

\begin{figure}[h]
    \centering
    \includegraphics[width=\linewidth]{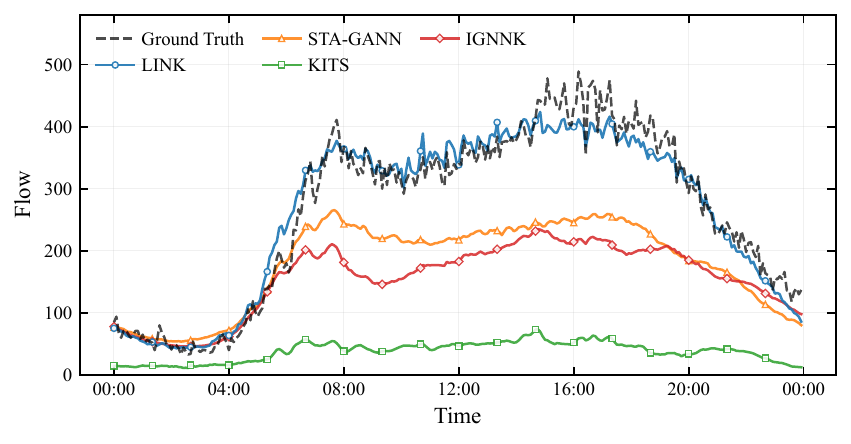}
    \caption{Kriging result with metadata-enhanced baselines}
    \label{fig:casewith}
\end{figure}

Fig.~\ref{fig:casewithout} presents the reconstruction results on a ramp node under the original setting, while Fig.~\ref{fig:casewith} shows the results with metadata-enhanced baselines on another mainline node in TraffiKrig-03. These two cases exhibit different traffic magnitudes, while their temporal patterns remain relatively similar.

Under the original setting, different baseline methods exhibit distinct reconstruction errors. KITS and IGNNK introduce pronounced morning and afternoon peaks that are absent in the ground truth, indicating that these methods fail to recover the target-specific temporal dynamics. In contrast, STA-GANN captures a temporal evolution that is more consistent with the ground truth, but substantially overestimates the traffic magnitude throughout the day. These results suggest that existing methods may capture general traffic variations or spatial correlations, yet still struggle to jointly preserve target-specific temporal patterns and traffic scales.

With the incorporation of road metadata, baseline methods demonstrate improved capability in capturing target-specific temporal dynamics. IGNNK and STA-GANN recover temporal patterns that are more consistent with the ground truth, indicating that road metadata provides useful contextual information for distinguishing different traffic behaviors among locations. However, these methods still exhibit considerable deviations in traffic magnitude, suggesting that auxiliary metadata alone is insufficient to accurately recover the absolute traffic scale. Among these methods, KITS shows the largest magnitude mismatch, where the reconstructed traffic volume substantially underestimates the ground truth. Overall, LINK achieves the closest reconstruction to the ground truth in both temporal evolution and traffic magnitude, demonstrating the effectiveness of the proposed propagation-based framework for recovering target-specific traffic states.

\section{Conclusion}\label{sec:conclusion}

In this work, we revisit traffic kriging from a propagation-centric perspective and propose LINK, a Lightweight and Inductive Node-to-node propagation rule learning paradigm for traffic Kriging. LINK explicitly models node-to-node traffic state transitions instead of relying solely on implicit neighborhood aggregation. In particular, it combines focal-specific propagation with a context-aware Spectral Propagation Network that performs node-to-node propagation in both the spectral and time domains. 
We further construct TraffiKrig to reveal the importance of road-specific metadata in traffic kriging and provide a unified metadata-enhanced evaluation protocol for systematic comparison across representative kriging methods.
Extensive experiments demonstrate that LINK achieves consistently strong reconstruction performance with high computational efficiency.  Compared with the strongest baseline on each subset, LINK reduces MAE by an average of \textbf{54.7\%}, while reducing training and inference time by \textbf{66.5\%} and \textbf{86.6\%}, respectively, relative to the fastest competing methods. Furthermore, zero-shot experiments provide additional evidence that LINK learns a transferable propagation rule that generalizes beyond the training region.

\clearpage 
\bibliographystyle{IEEEtran}
\bibliography{reference}

\end{document}